\documentclass{article}

\usepackage{microtype}
\usepackage{graphicx}
\usepackage{subfigure}
\usepackage{booktabs} 

\usepackage{hyperref}

\usepackage[accepted]{icml2023}

\usepackage{amsmath}
\usepackage{amssymb}
\usepackage{mathtools}
\usepackage{amsthm}

\usepackage[capitalize,noabbrev]{cleveref}

\theoremstyle{plain}

\theoremstyle{definition}

\theoremstyle{remark}

\usepackage[textsize=tiny]{todonotes}

\icmltitlerunning{Adaptive User-centered Neuro-symbolic Learning}

\begin{document}

\twocolumn[
\icmltitle{Adaptive User-centered Neuro-symbolic Learning for Multimodal Interaction with Autonomous Systems}




\begin{icmlauthorlist}
\icmlauthor{Amr Gomaa}{comp,yyy}
\icmlauthor{Michael Feld}{comp}
\end{icmlauthorlist}

\icmlaffiliation{yyy}{Saarland Informatics Campus, Saarland University, Saarbr{\"u}cken, Germany}
\icmlaffiliation{comp}{German Research Center for Artificial Intelligence (DFKI), Saarbr{\"u}cken, Germany}

\icmlcorrespondingauthor{Amr Gomaa}{amr.gomaa@dfki.de}

\icmlkeywords{Multimodal Interaction; Data Fusion; Adaptive Models; Personalization; Human-Centered Artificial Intelligence}

\vskip 0.3in
]



\printAffiliationsAndNotice{}  

\begin{abstract}
Recent advances in machine learning, particularly deep learning, have enabled autonomous systems to perceive and comprehend objects and their environments in a perceptual subsymbolic manner. These systems can now perform object detection, sensor data fusion, and language understanding tasks.
However, there is a growing need to enhance these systems to understand objects and their environments more conceptually and symbolically. It is essential to consider both the explicit teaching provided by humans (e.g., describing a situation or explaining how to act) and the implicit teaching obtained by observing human behavior (e.g., through the system's sensors) to achieve this level of powerful artificial intelligence.
Thus, the system must be designed with multimodal input and output capabilities to support implicit and explicit interaction models. In this position paper, we argue for considering both types of inputs, as well as human-in-the-loop and incremental learning techniques, for advancing the field of artificial intelligence and enabling autonomous systems to learn like humans. We propose several hypotheses and design guidelines and highlight a use case from related work to achieve this goal.
\end{abstract}

\begin{figure*}[h]
  \centering\includegraphics[width=0.9\linewidth]{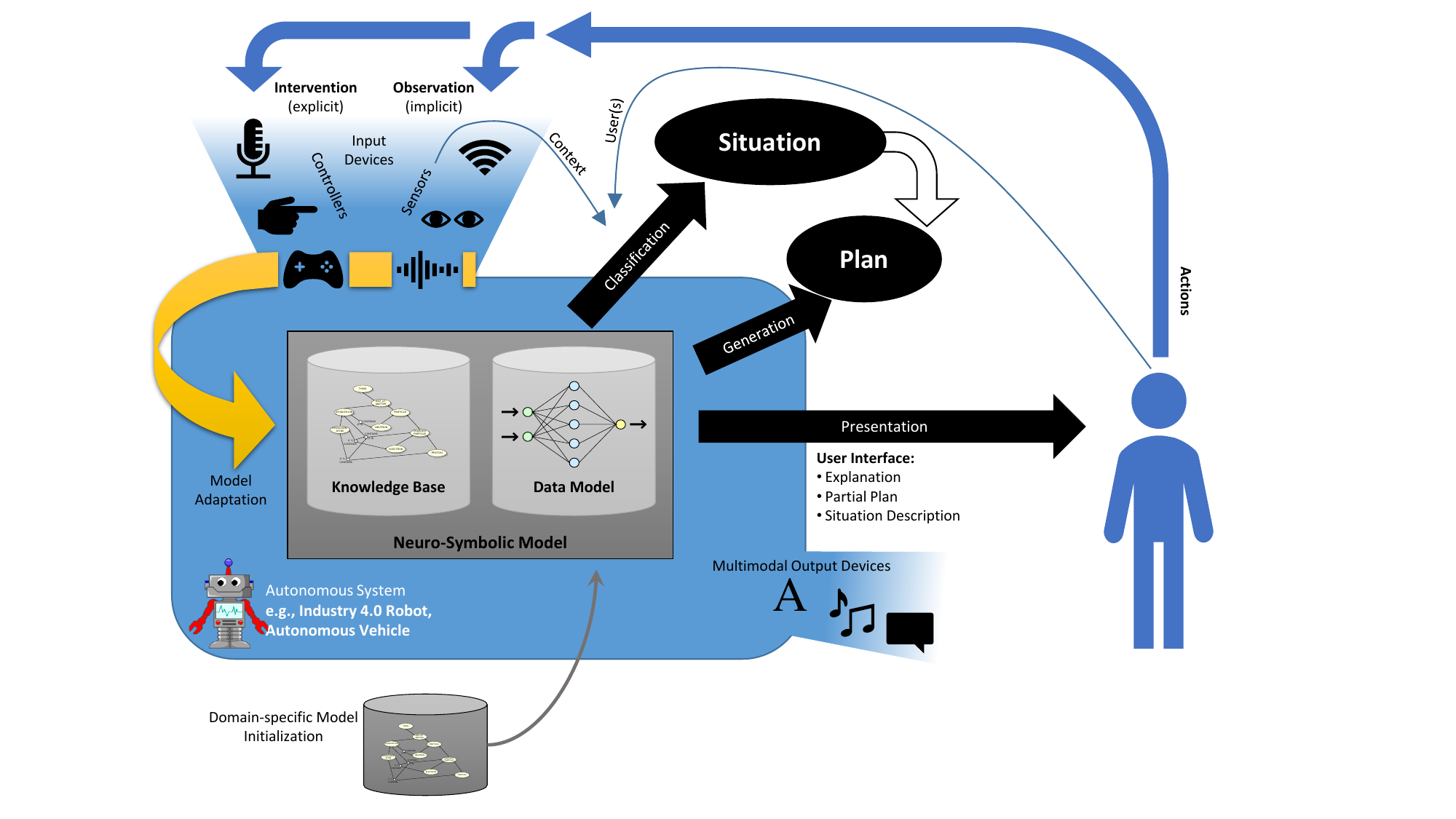}
  \caption{Overview of the envisioned user-centered neuro-symbolic human-in-the-loop learning system. The novice user and the learning agent (e.g., robot or autonomous vehicle) are in a continuous feedback loop, starting with the user's demonstrations, then judging the agent's output, and providing support through thorough feedback.}

  \label{fig:teaser}
\end{figure*}

\section{Introduction}

Human-centered artificial intelligence (HCAI) is an exciting new area of research that is attracting increasing attention from researchers of both artificial intelligence (AI) and human-computer interaction (HCI)~\cite{xu2019toward,nowak2018assessing,bryson2019society,shneiderman2020human}. Despite the significant progress that has been made in developing autonomous systems, these systems still rely heavily on human operators, whether local or remote, to step in and assist or take control in situations where the system is unable to proceed.
This highlights the need for HCAI techniques to promote trust, control, and reliability between users and machines~\cite{shneiderman2020human}. However, developing and implementing these concepts remains a challenging and complex task~\cite{nowak2018assessing}. As a result, there is still much room for improvement and further research in this field~\cite{bryson2019society}.

When it comes to multimodal interaction, a variety of approaches have been explored using early and late data fusion techniques~\cite{dong2009advances,karpov2018multimodal}. For example, researchers have studied hand and gaze fusion techniques for interacting with screen-based indoor systems~\cite{kim2017multi,zhang2015costs}. In the automotive domain, there has been a focus on controlling the vehicle and the infotainment system using touch-based approaches and multimodal combinations of hand gestures, gaze, and speech~\cite{sezgin2009multimodal,rumelin2013free,fujimura2013driver,Roider2017,feld2016combine}. Furthermore, it is crucial to consider the impact of cognitive load on driver ability to perform when using these interfaces and how it affects driving performance~\cite{Fuller2005TowardsBehaviour,Brown2020Ultrahapticons:Interfaces,Islam2020ALearning,Barua2020TowardsClassification,Solovey2014classifying}, which emphasizes the significance of personalization. Thus, we suggest that future work should focus on building autonomous systems that can learn and adapt to new situations, such as new classes, domains, or tasks~\cite{van2019threeincremental,von2019informed}. This will require shifting the focus from data-driven learning to interactive learning or human-in-the-loop learning, where the human plays a crucial role in supporting the system's learning process.

The proposed research concept focuses on developing adaptive and personalized approaches for human-in-the-loop learning that will enhance system performance and promote trust toward a reliable and controllable HCAI, as highlighted in~\autoref{fig:teaser}. More specifically, we highlight multiple methods and techniques for learning-based adapted models utilizing transfer-of-learning and propose some new aspects for continual learning for future work. Although these approaches apply to different domains, we focus on the automotive domain as an example of the rich work on driver personalization. More specifically, we demonstrate our suggestion on some of our previous work in the field of adaptive user interaction for the automotive domain~\cite{Gomaa2020,gomaa2021ml,gomaa2022adaptive,gomaa2022s,meiser2022vehicle,feld2019software}; however, the underlying learning techniques are valid for other domains as well.

\section{Background and related work}

Adaptive multimodal interaction combining speech, hand gestures, and gaze has been a topic of interest for the research community for the last 20 years in multiple domains, including robotics and automotive applications~\cite{rogers2000adaptive,hassel2005adaptation,janarthanam2014adaptive,manawadu2017multimodal,zhang2015costs,neverova2015moddrop,gnjatovic2012adaptive}. 
Despite the previously discussed significant advances in the adaptation of multimodal interaction, a personalized user-centered approach is still lacking. Thus, an important goal and factor in the proposed research work is user-specific personalization through incremental learning techniques~\cite{van2019threeincremental,gepperth2016incremental}.
As an example, in the automotive domain, researchers attempted multimodal fusion approaches for in-vehicle object selection in multiple works~\cite{roider2018see,Aftab2020,sezgin2009multimodal}. However, in-vehicle object referencing approaches do not generalize directly to outside-the-vehicle referencing, as the object's environment is static, limited, and in close proximity. Consequently, Moniri et al.~\cite{Moniri2012a} studied the single task of outside-the-vehicle referencing from the passenger seat using pointing, head pose, and eye gaze. Similarly, Aftab et al.~\cite{aftab2021multimodal} combined these modalities using a late fusion approach based on a neural network to reference objects from a stationary vehicle. While these approaches showed great promise, they still considered only a subsymbolic method for adaptation with a focus on data-driven approaches and did not consider user-specific behavior further.

Several approaches have proposed ways to insert human knowledge into neural networks as a way of initialization, to guide network refinement, and to extract symbolic information from the network~\cite{shavlik_combining_1994, von2019informed}. More recent attempts have tried to combine deep learning with knowledge bases in joint models (e.g., for construction and population)~\cite{ratnerAlexEtAl2018, adel2018deep}. Some work has focused on integrating neural networks with classical planning by mapping subsymbolic input to symbolic one, which automatic planners can use~\cite{asai2018classical}. Others have used Logic Tensor Networks to enable learning from noisy data in the presence of logical constraints by combining low-level features with high-level concepts~\cite{serafini2016logic, donadello_logic_2017}. Other approaches include psychologically inspired cognitive architectures having a goal-directed organizational hierarchy with parallel subsymbolic algorithms running at the lower levels and symbolic ones running serially at the higher levels~\cite{kelley_developing_2006}.
While subsymbolic learning methods, such as neural networks, have shown remarkable results in fields such as computer vision, NLP, and NLU, one problem they suffer from is a lack of explainability. On the other hand, while symbolic learning is ``legible'' by humans, it can lead to combinatorial growth that makes unfeasible solutions to complex problems~\cite{buker_hybrid_nodate}. When combining both types of learning, it could be possible to obtain advantages while overcoming the disadvantages. For example, a teacher might teach a robot how to tidy up a table full of bottles in different stages. In the first stage, the teacher might guide the robot's arm, showing it how to clear one bottle from the table (subsymbolic learning by example). In the next stage, when the basic movements have been acquired, supervised learning can continue through verbal instructions (symbolic learning by instruction)~\cite{grumbach_learning_1995}. 

\section{Research Questions and Hypotheses}

In line with the previous motivation and related work, the following research questions were developed to answer previous challenges from an abstract point of view while focusing on three factors \textit{Input features (i.e., Agent World View)}, \textit{Underlying design aspects (i.e., Multimodal interaction)}, and \textit{Learning method (i.e., Neuro-symbolic Adaptation and Continuous Learning)}. We envision these research questions as guidelines for future research on human-centered artificial intelligence.

\begin{figure*}[t]
	\begin{center}
		\includegraphics[width=0.75\linewidth]{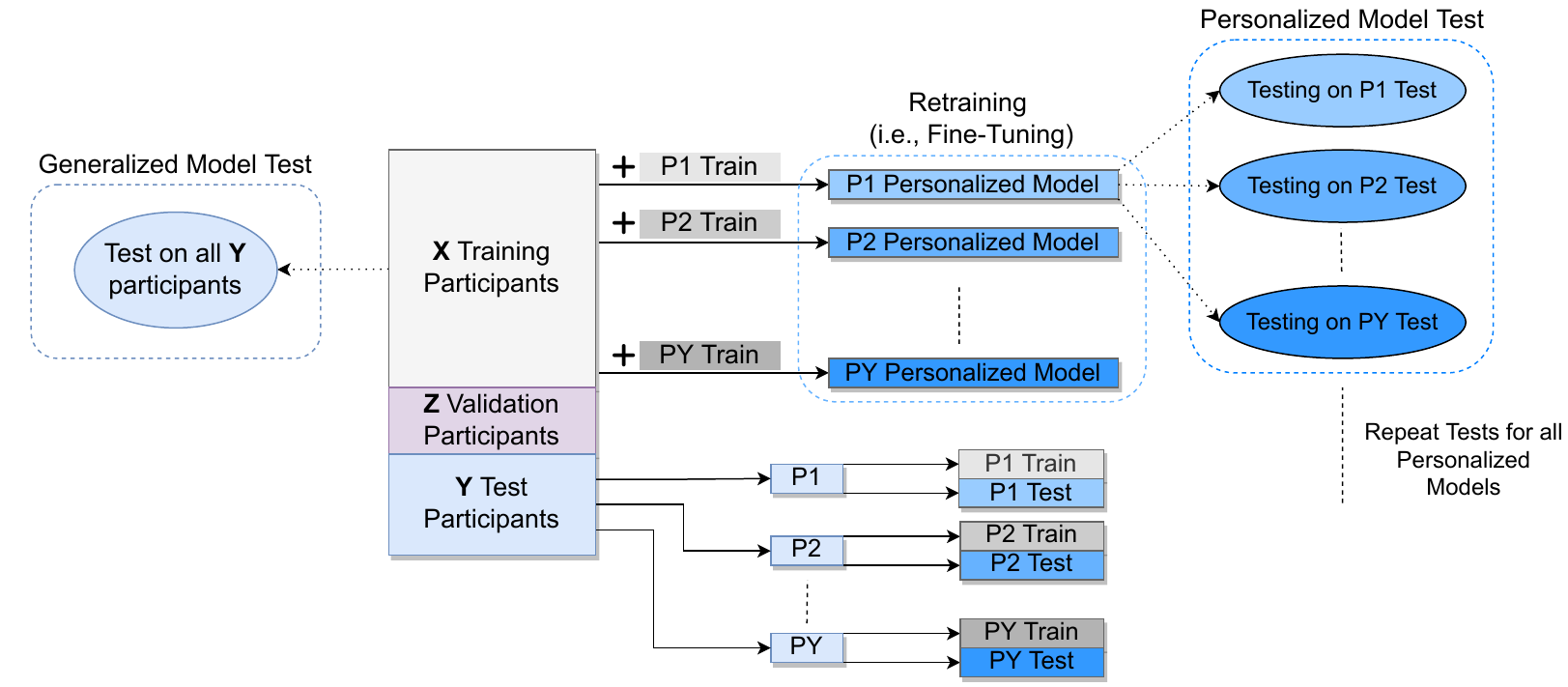}
	\end{center}
	\caption{Proposed approach for model adaptation to generate personalized models through transfer and incremental learning techniques from~\cite{gomaa2022adaptive}.}
	\label{fig:personalization}
\end{figure*}

\begin{itemize}
    \item \textbf{Agent World View (RQ1):} Which features of the agent (i.e., autonomous system) and the context (i.e., human behavior) can be used to detect and classify user interaction situations, and which devices are available to provide them efficiently (e.g., investigating user behavior as in~\cite{Gomaa2020})?

    Given the multitude of sensors available for an autonomous system, possibly dynamic and not permanently available, a specific question will be to select the right level of granularity and fusion at which it can be combined with symbolic knowledge. This involves merging the available context information, both from sensors and world knowledge, combined with implicit user input~\cite{knox2009interactively,cui2020empathic}, to characterize the situation in a structured way. For example, in an industry scenario, a worker's current task and the available robots would provide such input. In an autonomous vehicle scenario, knowledge about other passengers may help interpret the user's goals and possible interaction. Based on available plans and solutions, a system has to estimate the success of a particular solution.
    
    \item \textbf{Multimodal Interaction (RQ2):} What aspects of system and interface design can be utilized of the given modalities in terms of fusion techniques, temporal dependencies, and learning models to achieve optimal performance (e.g., reference detection as in~\cite{gomaa2021ml} and estimation of mental workload in~\cite{gomaa2022s,meiser2022vehicle})?

    To achieve an end-to-end multimodal fusion framework, it is vital to exhaustively investigate the interaction between the given modalities in terms of performance, timing, user behavior, and fusion techniques. While well-established, widely used data fusion approaches, such as late- and early-fusion approaches, are utilized here, more novel and empirical hybrid approaches should also be considered that combine heuristics with learning-based data fusion to achieve optimum performance. Additionally, there exists a timing dependency (e.g., modalities' relative onset) between the modalities that the system can exploit. Thus, the time frames can be analyzed separately with no connection, or a pattern could be learned from intra- (within the modality) and inter- (among the modalities) dependencies.
    
    \item \textbf{Neuro-symbolic Adaptation and Continual Learning (RQ3):} How can the system adapt to the performance of user-specific tasks~\cite{gomaa2021ml,meiser2022vehicle}? How can the system be designed to continuously gather feedback from the user (both implicitly and explicitly) to guarantee constant development and enhancement of the underlying algorithms? How would that affect the system's reliability and user trust?

    Adaptation can be achieved at the architecture level using incremental learning~\cite{gepperth2016incremental}. Transfer learning (i.e., naive fine tuning) faces several challenges such as forgetting previously learned information (i.e., catastrophic forgetting), ever-changing features (i.e., concept shift), and how fast a model should be adapted (i.e., stability-plasticity dilemma). Some solutions have been proposed for each of these challenges~\cite{schlimmer_incremental_1986, polikar_learn_2001,von2019informed}. For continuous learning, there is a focus on increasing the number of classes a neural network can predict, expanding datasets, and exploring the influence of update intervals and batch sizes used for adaptation~\cite{kading2016fine,van2019threeincremental}. To adapt an initial model to a different domain, we find suitable methods in the domain of incremental learning~\cite{long2017deep, jie2011multiclass,de2021continual}.


\end{itemize}

\section{Proposed Methodology}

The proposed method follows the previously mentioned research questions to propose a multistage approach to reach an adaptive neuro-symbolic autonomous system with continuous user feedback (as seen in~\autoref{fig:teaser}).

The first stage of the proposed plan is to understand the variances in driver behavior when performing the multimodal referencing task as in~\cite{Gomaa2020,gomaa2022s,rumelin2013free}. As an example, in the automotive domain, drivers perform different multimodal gestures to control the vehicle and query surrounding objects. These individual differences could be exploited by the system for personalization and adaptation through a user-centered design approach. Drivers could be clustered based on single-modality performance, and a switching mechanism could be applied within the overall system~\cite{Gomaa2020} to maximize overall performance (i.e., turn gaze detection off for user accompanied by wandering behavior of the eye and thus low accuracy of gaze detection). Furthermore, understanding the mental workload patterns of users could be exploited by the system and also to enhance its performance through model adaptation and personalization~\cite{gomaa2022s}.
The second stage would be creating an end-to-end learning-based multimodal fusion framework through constant and exhaustive monitoring of the users through system sensors. This is an initial step to automate the previously mentioned heuristics by the system~\cite{gomaa2021ml} using hybrid learning where a pattern could be learned from intra- (within the modality) and inter- (among the modalities) dependencies. However, adaptation is an inherently continuous paradigm; thus, it is considered an ongoing process along the user observations and the multimodal fusion stages in drivers' categorization (i.e., clustering) and hybrid fusion approaches, respectively. Although model adaptation, in the previous context, is one alternative to the one-model-fits-all approach, it still groups users in a particular model (i.e., cluster), constituting a many-models-fits-all approach. However, a more personalized approach would utilize transfer-of-learning and incremental learning techniques to eventually reach a single model or continuously adapting model per user.~\autoref{fig:personalization} shows an approach to achieving these personalized models through incremental learning techniques. The data set is initially divided into training, validation, and test sets as in traditional learning approaches. The model is trained on $\textbf{X}$ participants' data while the hyperparameters are chosen and validated on $\textbf{Z}$ participants' data, and the final model is tested on $\textbf{Y}$ participants' data. On the other hand, for the adaptation approach, each participant's data from the $\textbf{Y}$ test set are further split (e.g., equally) into subtrain and subtest sets where the model is retrained and fine-tuned on the user-specific training data to produce personalized model weights that are optimized for this user. To assess the effect of this approach, the personalized model is tested on the same participant sub-test data and compared against other participants' sub-test data.

     \begin{figure}[t]
         \centering
         \includegraphics[width=0.8\linewidth]{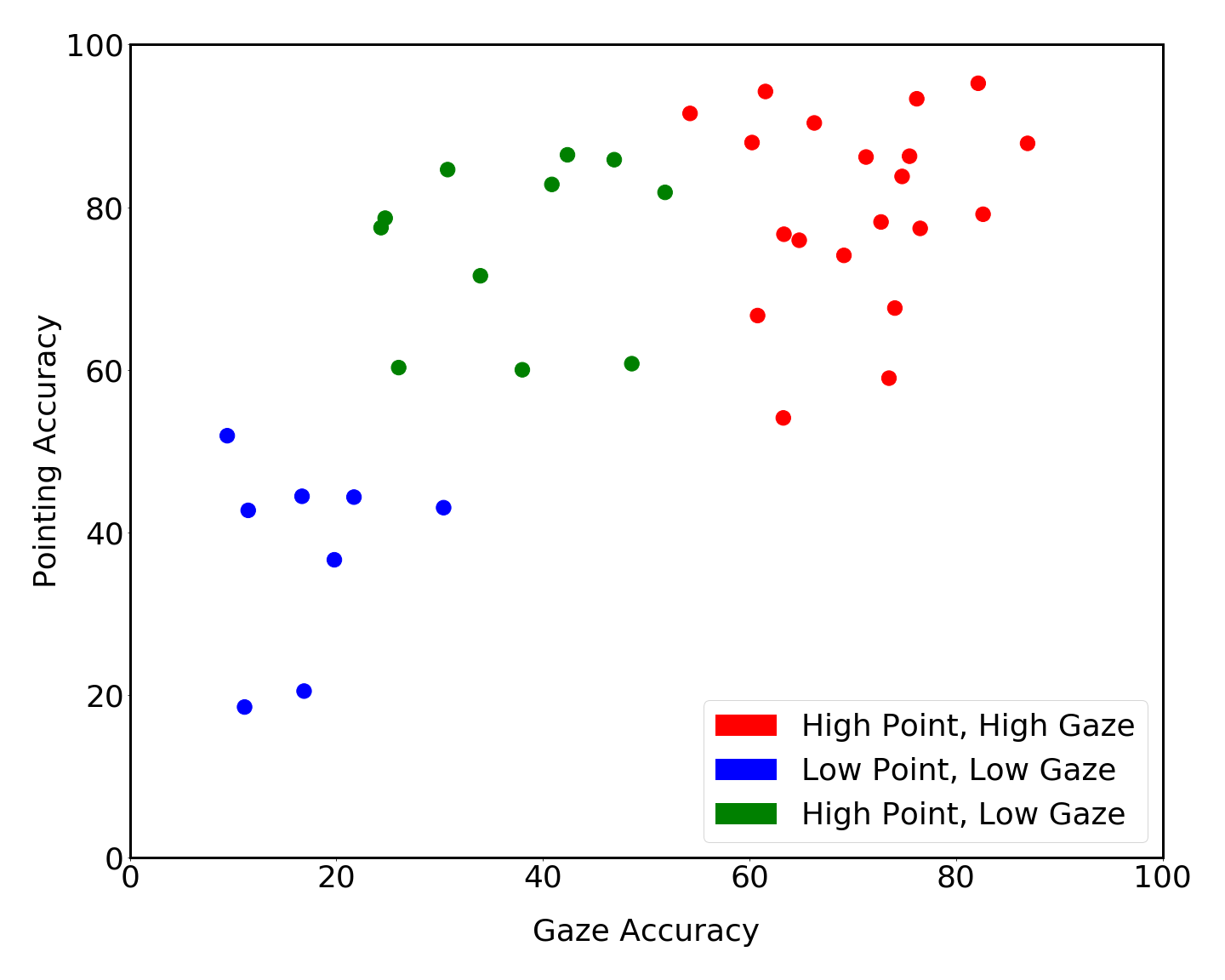}
         \caption{Clustering drivers' pointing and gaze behavior based on the system's perceived performance (i.e., referencing accuracy) from~\cite{Gomaa2020}.}
         \label{fig:clustering}
     \end{figure}
    
     \begin{figure}[t]
         \centering
         \includegraphics[width=0.8\linewidth]{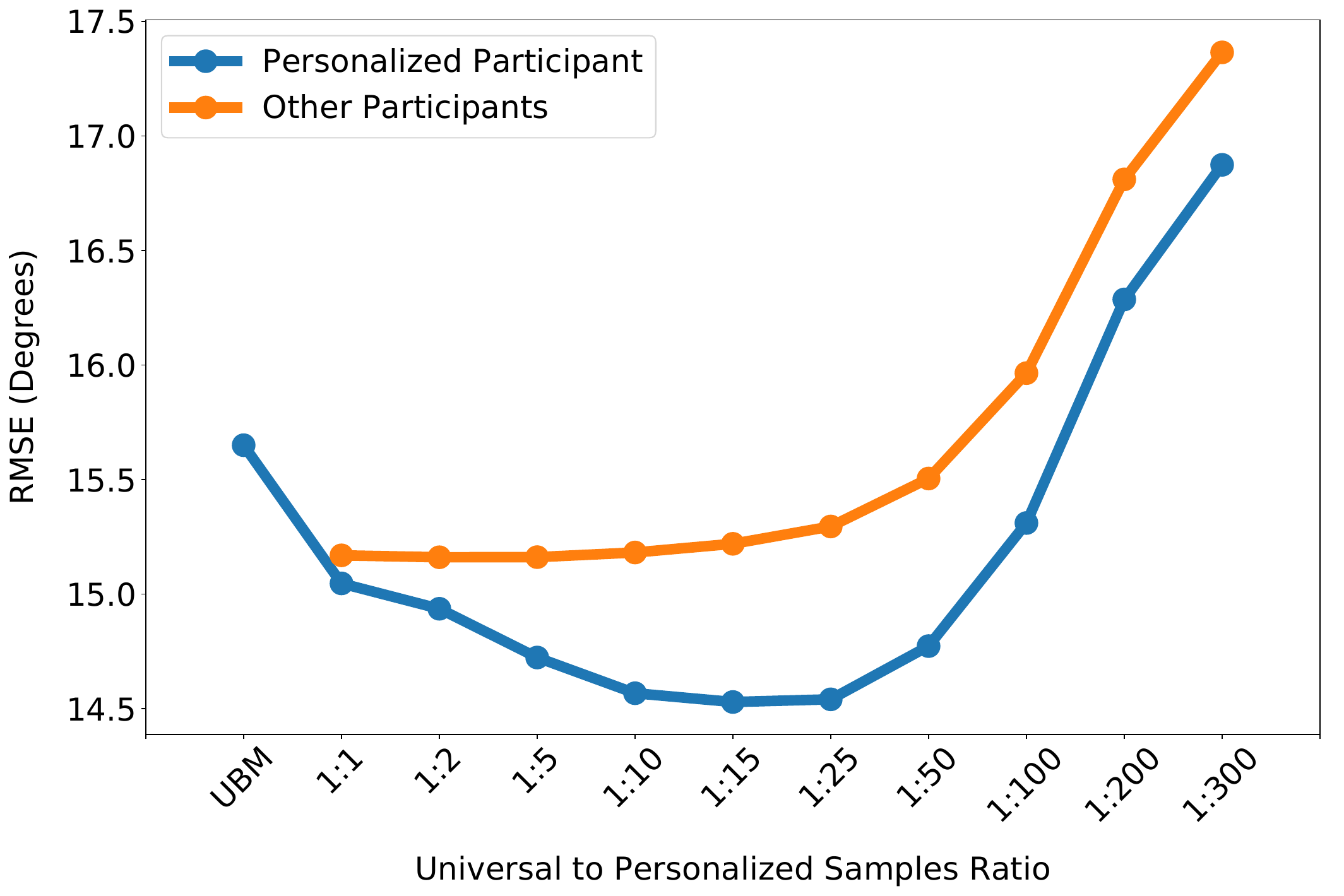}
         \caption{RMSE results comparing different personalized models among themselves (based on the sample weight of the user-specific data) and against the UBM (i.e., generalized model) from~\cite{gomaa2021ml}.}
         \label{fig:personalizationsampleweight}
     \end{figure}

While the previous approaches optimize system performance based on current individual behavior, this behavior could change over time due to situational, emotional, or mental load variations and learning effects. Thus, a continuous learning approach is considered where the user can give feedback to the system implicitly (e.g., via dissatisfied looks or grunting as visual or auditory cues) or explicitly (e.g., repeating the given voice command). To achieve this goal, the study and data collection phase should include different variations in the situational and mental state for internal and external validity. Finally, situation-adapting learning techniques could be further utilized in this context, such as graph classification and node selection (e.g., Relational Graph Neural Networks~\cite{jing2020relational}), learning from the driver's behavior (e.g., Efficient Learning from Demonstrations~\cite{li2022efficient}), and learning from the driver's feedback (e.g., Implicit Human Feedback Learner~\cite{cui2020empathic}).

Since the main focus of this work is on adaptation and user-specific personalization,~\autoref{fig:clustering} and~\autoref{fig:personalizationsampleweight} show examples of related work results focusing on the adaptation aspect of~\cite{Gomaa2020}. Specifically,~\autoref{fig:clustering} shows how drivers' referencing actions could be clustered based on pointing and gaze modality performance separately; then, each cluster is trained independently. Thus, each cluster model weight would be adapted to the cluster pointing- and gaze-specific accuracy. This resembles the hybrid-fusion approach discussed earlier. Similarly,~\autoref{fig:personalizationsampleweight} highlights the results of the incremental learning personalization approach previously discussed in~\cite{gomaa2021ml}. It compares the personalized model subtest data against the average of the other non-personalized subtest data using the Root Mean Square Error (RMSE) metric. The figure also highlights further enhancement of this personalization approach; it was noticed that adding the subtrain data of the personalized participant to the existing generalized model (also called Universal Background Model (UBM)) data with a 1:1 ratio is not the optimum solution due to its insignificant contribution size. Therefore, personalized participant subtrain data was emphasized (e.g., repeating the data multiple times), and its ratio increased for the training data $\textbf{X}$ with a ratio of 1:2, 1:5, etc. until the optimum sample weight could be determined. While we focus precisely on these results for the referencing task, the methodology applies to any regression problem with a similar setup. Thus, it can be generalized to multiple sensors and multimodal platforms.

\section{Conclusion}

Although designing user-specific interfaces is a complex and multifaceted process involving various considerations that this work cannot entirely describe, our position paper examines several essential aspects to facilitate this design process. Specifically, we discuss adapting learning models, including incremental and transfer learning, to enable personalized interaction with the system. This work also emphasizes the importance of system engineering considerations, such as real-time processing and system robustness, to ensure that user-specific interfaces are reliable and trustworthy.
This paper highlights important considerations for future studies focused on human-centered artificial intelligence and trustworthy interfaces. In particular, we emphasize the importance of continuous learning and hybrid learning approaches to enable user-centered design that enhances the user experience. By following these guidelines, researchers can develop personalized and adaptive interfaces that respond to individual users' needs and behaviors, ultimately improving their satisfaction and engagement with the system. Furthermore, future research in this area should focus on developing frameworks and methodologies to assess the effectiveness of user-specific interfaces and explore the ethical and societal implications of these technologies.

\section*{Acknowledgements}

This work is partially funded by the German Ministry of Education and Research (BMBF) under the TeachTAM project (Grant Number: 01IS17043) and the CAMELOT project (Grant Number: 01IW20008).


\bibliography{sample-base}
\bibliographystyle{icml2023}



\end{document}